\documentclass{article} 
\usepackage{iclr2022_conference,times}

\usepackage{hyperref}
\usepackage{url}
\usepackage{adjustbox}
\usepackage{array}
\usepackage{pifont}
\usepackage{pgfplots}
\pgfplotsset{width=10cm,compat=1.9}

\usepgfplotslibrary{external}
\title{Vision Transformers in 2022:\\ An Update on Tiny ImageNet}

\author{Ethan M. Huynh\\
University of California, San Diego\\
\texttt{e2huynh@ucsd.edu} \\
}

\iclrfinalcopy 
\begin{document}
\maketitle

\begin{abstract}
The recent advances in image transformers have shown impressive results and have largely closed the gap between traditional CNN architectures. The standard procedure is to train on large datasets like ImageNet-21k and then finetune on ImageNet-1k. After finetuning, researches will often consider the transfer learning performance on smaller datasets such as CIFAR-10/100 but have left out Tiny ImageNet. This paper offers an update on vision transformers' performance on Tiny ImageNet. I include Vision Transformer (\textbf{ViT}) , Data Efficient Image Transformer (\textbf{DeiT}), Class Attention in Image Transformer (\textbf{CaiT}), and \textbf{Swin Transformers}. In addition, \textbf{Swin Transformers} beats the current state-of-the-art result with a validation accuracy of \textbf{91.35\%}. Code is available here: \url{https://github.com/ehuynh1106/TinyImageNet-Transformers}
\end{abstract}

\section{Introduction}

The ViT paper \citep{vit} showed that transformers can be applied to image classification tasks. However ViT was pretrained on the JFT-300M dataset \citep{jft300m}, Google's internal dataset of 300 million images. Thus the issue of training efficiency and data availability was apparent. DeiT \citep{deit} was a response to that and showed that a way to alleviate the data-hungry nature of transformers was with a rigorous training schedule and knowledge distillation. As such, it became possible to train a vision transformer using ImageNet-21k \citep{imagenet21k} and further finetuning on ImageNet-1k \citep{imagenet1k}. Subsequent image transformers, like CaiT \citep{cait} and Swin \citep{swin}, closely follow the blueprint laid out by DeiT. 

In addition to ImageNet-1k, these studies perform transfer learning tests on CIFAR-10 and CIFAR-100 \citep{cifar}. However, every paper has failed to include Tiny ImageNet \citep{tinyimagenet}. Tiny ImageNet is a subset of ImageNet-1k with 100,000 images and 200 classes that was first introduced in a computer vision course at Stanford. Since its inception, few papers have used this dataset in their benchmarks.

That being said, a study has been done by \cite{vitstudy} where they propose modifications to vision transformers to improve the accuracy training from scratch on Tiny ImageNet. But in reality, transfer learning is a much more common and stronger technique when it comes to accuracy. As such, there is no modern research done to evaluate vision transformers on Tiny ImageNet. This paper address that gap and will report the accuracy of \textbf{ViT}, \textbf{DeiT}, \textbf{CaiT}, and \textbf{Swin transformer} using a training regiment similar to DeiT's.

\section{Experimental setting}
\label{experimentalsetting}

All vision transformers are taken from the \texttt{timm} library \citep{rw2019timm}. I trained each model using a Nvidia RTX 3070 (8GB memory) and an 8-core CPU. The size of the models were chosen based on being within the neighborhood of 10 to 60 minutes to train per epoch. As a refresher, I report the accuracies of the transformers on ImageNet-1k in \hyperref[table1]{\bf Table 1}.

{\renewcommand{\arraystretch}{1.07}%
\begin{table}[h]
\begin{center}
\begin{tabular}{l||c|c|c}
\bf Model &\bf ImageNet-1k &\bf CIFAR-100 &\bf CIFAR-10\\
\hline
ViT-L/16   &87.08    &\bf94.04    &99.38 \\
DeiT-B/16-D   &85.43    &91.40    &99.20 \\
CaiT-M/36   &86.05  &93.10    &\bf99.40 \\
Swin-L/4    &\bf87.15  &-  &-\\
\hline
\end{tabular}
\caption{Results of ViT, DeiT, CaiT, and Swin on ImageNet-1k, CIFAR-100, and CIFAR-10. All models are finetuned on 384x384 resolution. The numbers for ImageNet-1k are taken from the \texttt{timm} library and the rest are from the original papers. -D signifies distillation. The authors of the Swin transformer did not report the accuracy on CIFAR-10 and CIFAR-100.}
\label{table1}
\end{center}
\end{table}}

\subsection{Data augmentation}

The data augmentation techniques used mainly reflect that of DeiT's. I use Mixup \citep{mixup} and Cutmix \citep{cutmix} with a probability of 0.8 and 1.0 respectively, and Random Erasing \citep{randerase} with a 0.25 probability. I train using the full image resized to 384x384 resolution with bicubic interpolation in both training and testing, as suggested by \citet{deit}.

\subsection{Regularization \& Optimizer}

For regularization, I employ label smoothing with an $\epsilon$ of 0.1 and a stochastic depth of 0.1

I train each model for 30 epochs, using 128 batch size. With the image resolution being 384x384, 8 gigabytes of video memory is not enough to load both the model and batch into GPU memory. As such, gradient accumulation was required to train with a 128 batch size.

The optimizer of choice is AdamW at an initial learning rate of 10$^{-3}$ with cosine decay and weight decay of 0.05.

\section{Results}

{\renewcommand{\arraystretch}{1.1}%
\begin{table}[h]
\begin{center}
\begin{tabular}{l||c|c|c}
\bf Model  &\bf \begin{tabular}[c]{@{}c@{}}Tiny\\ImageNet\end{tabular} &\bf \#params &\bf FLOPs
\\ \hline
ViT-L/16                 &86.43    &304M    &190.7B\\
CaiT-S/36                &86.74  &68M    &48.0B\\
DeiT-B/16-D   &87.29    &87M    &55.5B\\
Swin-L/4                 &\bf91.35  &196M  &103.9B \\
\hline
\end{tabular}
\caption{Analysis of ViT, DeiT, CaiT, and Swin accuracy and training efficiency on Tiny ImageNet. I report the highest validation accuracy obtained during training. Throughput is measured with a batch size of 32. This is largest possible batch size that can fit on all models.} 
\label{table2}
\end{center}
\end{table}}

\paragraph{Swin} The large Swin transformer achieves state-of-the-art accuracy of \textbf{91.35\%}, beating the previous by 0.33\%. Applying a window to multi-headed self-attention (MSA) and a shifted window to MSA proves to be effective. Swin continues to impress among the vision transformers.

\paragraph{ViT} The ancestor of vision transformers, ViT, falls behind its advancements. The \texttt{timm} library reports that ViT outperforms DeiT by a significant margin on ImageNet-1k and \cite{deit} reports that ViT-L performs worse then a distilled DeiT-B so perhaps there is an advancement in training in the \texttt{timm} library I'm not aware of. 

\paragraph{DeiT} The base distilled DeiT achieves a respectable accuracy of \textbf{87.29\%} while training the fastest by a large margin. The power of knowledge distillation is evident as it performs better then ViT-L and trains the fastest by a large margin.

\paragraph{CaiT} CaiT has an accuracy of \textbf{86.74\%}. Considering that CaiT-S/36 is based off a DeiT-S, the accuracy is expected. A CaiT-M/36 model would likely surpass the DeiT model but CaiT took the longest to train despite the parameter count and number of FLOPs.

\subsection{Parameters and FLOPs are not created equally}

CaiT shows that parameter count and FLOPs are not indicative of model efficiency. Despite CaiT-S/36 having the lowest parameter count and FLOPs, it reports the lowest throughput and trains the slowest, see \hyperref[table3]{\bf Table 3}. Instead, there is a trend with layer count and throughput. The size of the embedding is another thing to consider but the number of layers seems to be the main determining factor considering CaiT's small embedding size.

{\renewcommand{\arraystretch}{1.1}%
\begin{table}[h]
\begin{center}
\begin{tabular}{l||c|c|c|c|c}
\bf Model  &\bf \#layers &\bf \begin{tabular}[c]{@{}c@{}}Embedding\\Size\end{tabular} &\bf \#params &\bf FLOPs &\bf \begin{tabular}[c]{@{}c@{}}Throughput\\(images/sec)\end{tabular}
\\ \hline
ViT-L/16                &24 &1024 &304M    &190.7B  &31.5\\
CaiT-S/36               &36 &368 &\bf 68M    &\bf 48.0B  &24.0\\
DeiT-B/16-D    &\bf 12 &768  &87M    &55.5B &\bf 83.1\\
Swin-L/4                &18 &\bf 192 &196M  &103.9B   &36.0\\
\hline
\end{tabular}
\caption{Comparison of various model size metrics with throughput.}
\label{table3}
\end{center}
\end{table}}

\section{Tuning the training procedure}

This section describes the experiments I did with the training procedure setup to reach the final setup described in \hyperref[experimentalsetting]{\bf Section 2} and achieve a final validation accuracy of: \textbf{91.35\%}.

\subsection{Hyperparameters}

\paragraph{Optimizers} For AdamW, I tried a combination of learning rates and weight decay in the range of [$3.10^{-3}$, $10^{-3}$, $7.10^{-4}$, $5.10^{-4}$] and [0.2, 0.05, 0.01] respectively. I found the setting of $10^{-3}$ learning rate and 0.05 weight decay to work the best.

Perturbed optimizers like SAM \citep{sam}, ASAM \citep{asam}, and PUGD \citep{pugd} were also considered. Initial testing showed that SAM, ASAM, and PUGD increased the training time of an epoch by 85\% while only having accuracies around 60-70\% for the first 5 epochs. In comparison, AdamW has 89\% accuracy after the first epoch. As a result, I decided not to train with these optimizers.

SGD was also tested with a learning rate of $10^{-2}$, weight decay of $10^{-5}$, and momentum of 0.9, with and without nesterov momentum.

\begin{figure}
    \centering
    \includegraphics{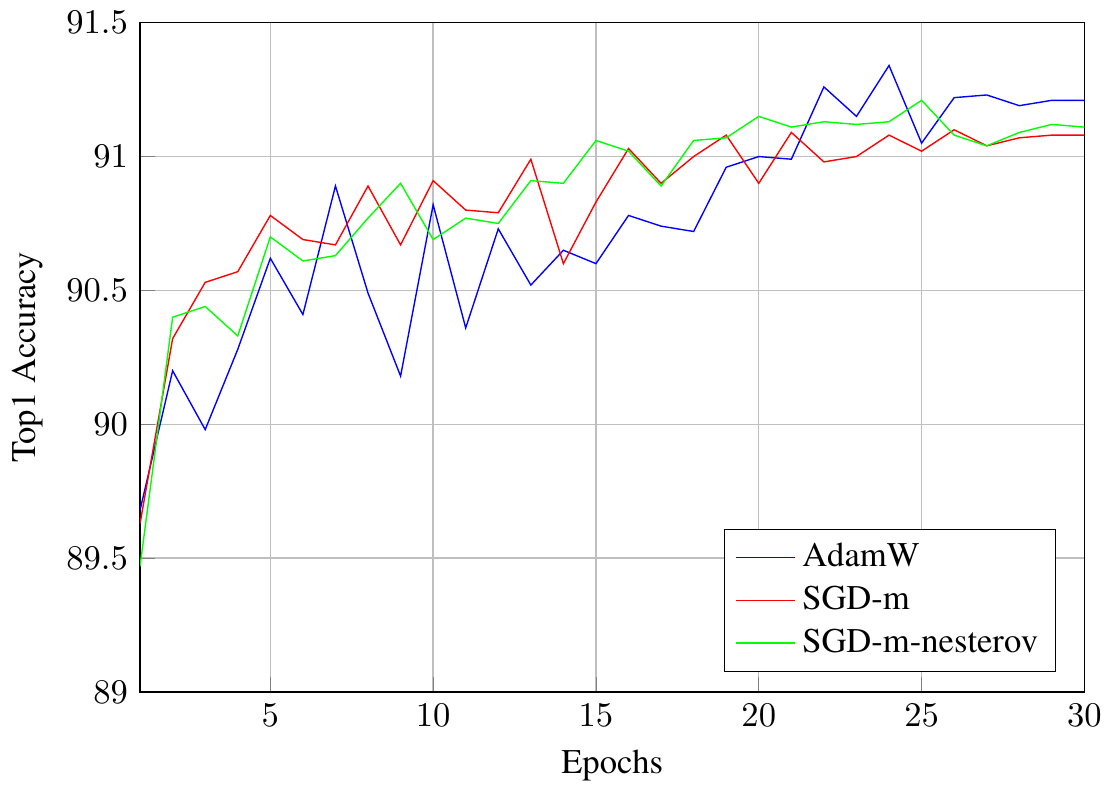}
    \caption{Convergence of AdamW vs SGD}
    \label{figure1}
\end{figure}

\hyperref[figure1]{\bf Figure 1} shows that SGD converges faster then AdamW, likely due to the fact that I use a higher learning rate for SGD. On the other hand, AdamW is more variable then SGD perhaps because of the adaptive nature of AdamW. Either way, SGD falls short of AdamW in terms of accuracy but nesterov momentum performs better than vanilla momentum (91.21\% vs 91.1\%).


\subsection{Ablation Study}

This section describes the effects of removing various data augmentation and regularization techniques as well as trying different ones to determine what the optimal training setup is for Swin on Tiny ImageNet.

I define the default training setup to be: Random Augment \citep{randaugment}, Random Erasing, Mixup, CutMix, Stochastic Depth, and Label Smoothing. Experimentation was done by simply removing one factor at a time and comparing their performance with the default setting.

In addition to those techniques, there were 3 more techniques I experimented with: AutoAugment \citep{autoaug}, image cropping, and Model EMA. 

AutoAugment was a widely-used data augmentation technique before the introduction of RandAugment. Research has since favored the simplicity of RandAugment but AutoAugment is still comparable to RandAugment and is still considered when designing training schemes. 

There are 2 types of cropping I tried, Random Resized Crop (RRC) and Simple Random Crop (SRC) \citep{deit3}. RRC is the standard method of cropping where the crop is just a random portion of the image resized to a given resolution. SRC is a square crop along the x-axis which maintains the aspect ratio while having a bigger fraction of the image. The main goal of SRC is that it is more likely to have the label in the image.

Model EMA has shown that in some cases it boosts accuracy and in some cases it hurts accuracy. \citet{swin} reports that Model EMA does not help but it is still worth trying on a dataset-by-dataset basis.

{\renewcommand{\arraystretch}{1.1}%
\begin{table}[h]
\begin{center}
\begin{tabular}{p{7cm}||c}
\bf Change &\bf Accuracy
\\ \hline
Default                                 &91.34\\
\hline
Remove RandAugment                     &\bf 91.35\\
Remove Rand-Erasing                     &91.28\\
Add Simple Random Crop                  &91.26\\
Remove Stochastic Depth                 &91.25\\
Remove Label Smoothing                  &91.25\\
Replace RA with AutoAugment             &91.25\\
Remove Mixup                            &91.11\\
Add Random Resized Crop                 &91.06\\
Remove CutMix                           &91.04\\
Add Model EMA                           &90.83\\
\hline
\end{tabular}
\caption{Comparison of the effects of various data augmentation and regularization techniques. RA stands for Rand-Augment.}
\label{table4}
\end{center}
\end{table}}

\subsection{Ablation Results}

I provide \hyperref[table4]{\bf Table 4} for detailed results. RandAugment was the only data augmentation technique in the default setting to technically decrease training accuracy. Without RandAugment, the model trains to a 91.35\% accuracy which is within margin of error of the default setting. However, the training patterns exhibited were attractive. The model was able to maintain its top accuracy for multiple epochs whereas the default setting peaks at 91.34\% and plateaus at around 91.20\%. For that reason, I decided to remove RandAugment. The other augmentation and regularization methods proved to be beneficial towards learning.

AutoAugment and Model EMA show a decrease in accuracy. \citet{swin} mentions that Model EMA does not increase performance, so my findings fall in line with the original paper. Whereas AutoAugment, I did not put in the time to try to optimize the parameters so perhaps with some tuning it could be useful.

Both cropping methods aren't effective in this environment. It's likely due to the fact that some form of cropping was utilized during ImageNet-21k pretraining and ImageNet-1k finetuning. At that point, cropping is not required as the model has already been strictly regularized.

As such, the final training setup is as follows:

\begin{itemize}
    \item AdamW with $10^{-3}$ learning rate and 0.05 weight decay
    \item 30 epochs, cosine decay, and 128 batch size
    \item Mixup
    \item CutMix
    \item Random Erasing
    \item Stochastic Depth of 0.1
    \item Label smoothing of 0.1
\end{itemize}

\section{Conclusion}

This paper has shown that vision transformers transfer well onto Tiny ImageNet which makes sense as it is a subset of ImageNet-1k. Two standout architectures are DeiT and Swin. DeiT reports a respectable accuracy while training the fastest by a significant margin and Swin achieves state-of-the-art accuracy, beating the previous by 0.33\%. Future work could be done on even more vision transformers. SwinV2 \citep{swinv2} improves on Swin and scales up Swin using self-supervised learning techniques and post normalization among other things. Another is MiniViT \citep{MiniViT}, who uses a combination of weight sharing and weight distillation to drastically reduce the parameter count and increase accuracy of vision transformers. 

\bibliographystyle{iclr2022_conference}
\bibliography{iclr2022_conference}

\appendix

\end{document}